\pdfoutput=1

\documentclass[11pt]{article}

\usepackage[preprint]{acl}
\usepackage{amsmath} 
\usepackage{amsfonts}
\usepackage{multirow}
\usepackage{url}
\usepackage{algorithm}
\usepackage{algorithmic}
\usepackage{times}
\usepackage{latexsym}
\usepackage{mathtools, nccmath}
\usepackage[T1]{fontenc}

\usepackage[utf8]{inputenc}

\usepackage{microtype}

\usepackage{inconsolata}
\usepackage{graphicx}
\usepackage{gnuplottex}
\usepackage{comment}

\usepackage{caption}
\usepackage{subcaption}
\usepackage{flushend}

\usepackage{xspace}
\newcommand{\name}{\texttt{E-Sparse}\xspace}
\usepackage{bbding}
\usepackage{booktabs}

\newcommand*\samethanks[1][\value{footnote}]{\footnotemark[#1]}
\newcommand*\authornote[1][\value{footnote}]{This work was done when Lin Niu was an intern at Tencent.}

\title{E-Sparse: Boosting the Large Language Model Inference through Entropy-based N:M Sparsity}

\author{Yun Li\thanks{\hspace{.06in}Equal contribution}  \\
  MLPD, Tencent \\
  \texttt{charlesyli@tencent.com} \\ \And
  Lin Niu\samethanks \thanks{\hspace{.06in}This work was done when Lin Niu was an intern at Tencent.} \\
  Huazhong University of Science \\ and Technology \\
  \texttt{linniu@hust.edu.cn} \\ \And
  Xipeng Zhang \\
  MLPD, Tencent \\
  \texttt{xipengzhang@tencent.com} \\ \AND
  Kai Liu \\
  MLPD, Tencent \\
   \texttt{raccoonliu@tencent.com} \\ \And
  Jianchen Zhu \\
  MLPD, Tencent \\
   \texttt{dickzhu@tencent.com} \\ \And
  Zhanhui Kang \\
  MLPD, Tencent \\
   \texttt{kegokang@tencent.com} \\
  }

\begin{document}

\maketitle

\begin{abstract}
Traditional pruning methods are known to be challenging to work in Large Language Models for Generative AI because of their unaffordable training process and large computational demands.
For the first time, we introduce the information entropy of hidden state features into a pruning metric design, namely \name, to improve the accuracy of N:M sparsity on LLMs.
\name employs the information richness to leverage the channel importance, and further incorporates several novel techniques to put it into effect:
(1) it introduces information entropy to enhance the significance of parameter weights and input feature norms as a novel pruning metric, and performs N:M sparsity without modifying the remaining weights.
(2) it designs global naive shuffle and local block shuffle to quickly optimize the information distribution and adequately cope with the impact of N:M sparsity on LLMs' accuracy.
\name is implemented as a Sparse-GEMM on FasterTransformer and runs on NVIDIA Ampere GPUs.
Extensive experiments on the LLaMA family and OPT models show that \name can significantly speed up the model inference over the dense model (up to 1.53$\times$) and obtain significant memory saving (up to 43.52\%), with acceptable accuracy loss.
\end{abstract}


\section{Introduction}

\begin{figure}[t]
\centering
\subfloat[Evaluate the input features from both cross-channel and intra-channel dimensions.]{\includegraphics[width=1.0\linewidth]{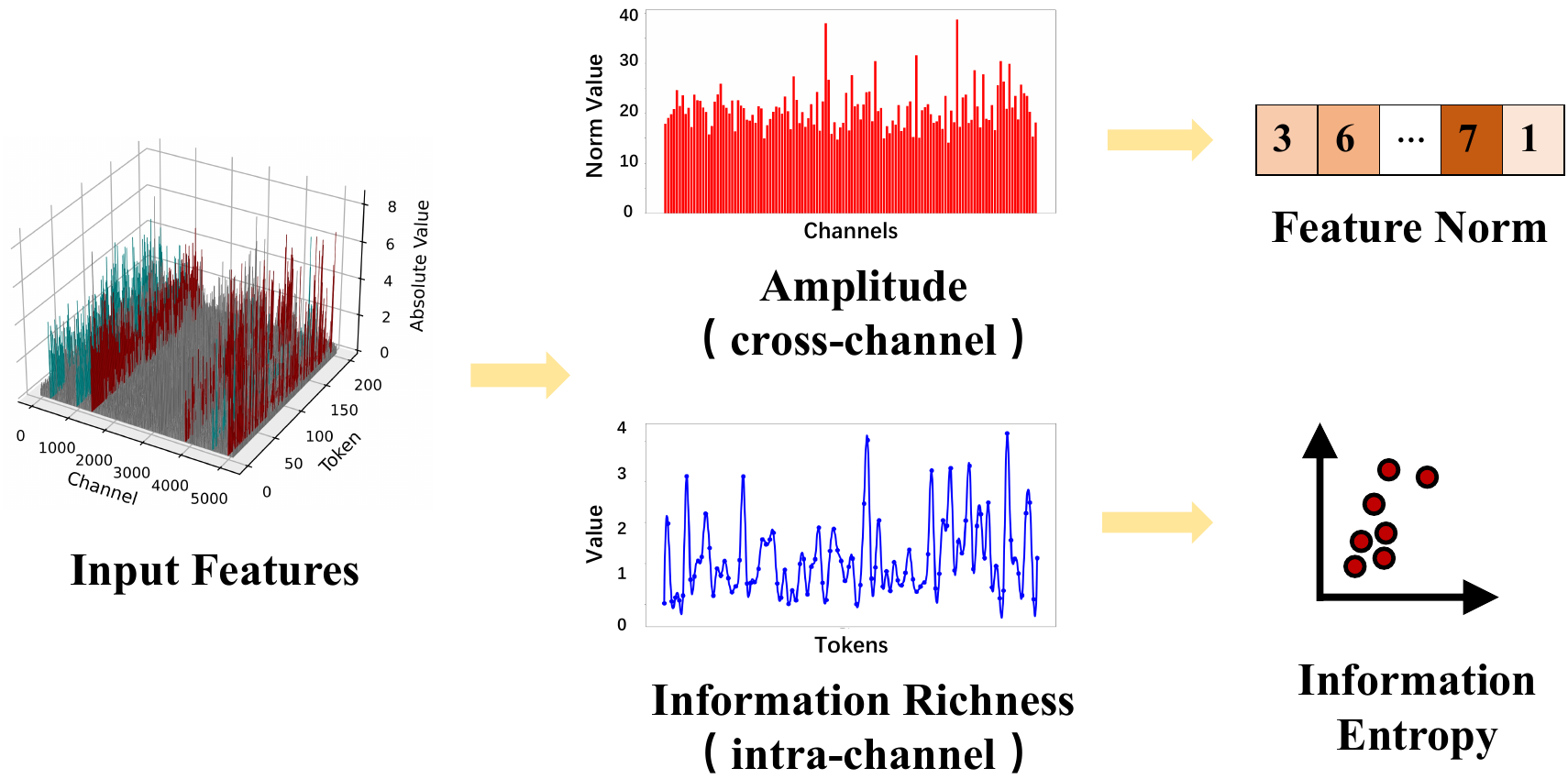} \label{fig:metric_1}} \\
\vspace{0.5cm}
\subfloat[The entropy-based sparsity metric of \name.]{\includegraphics[width=1.0\linewidth]{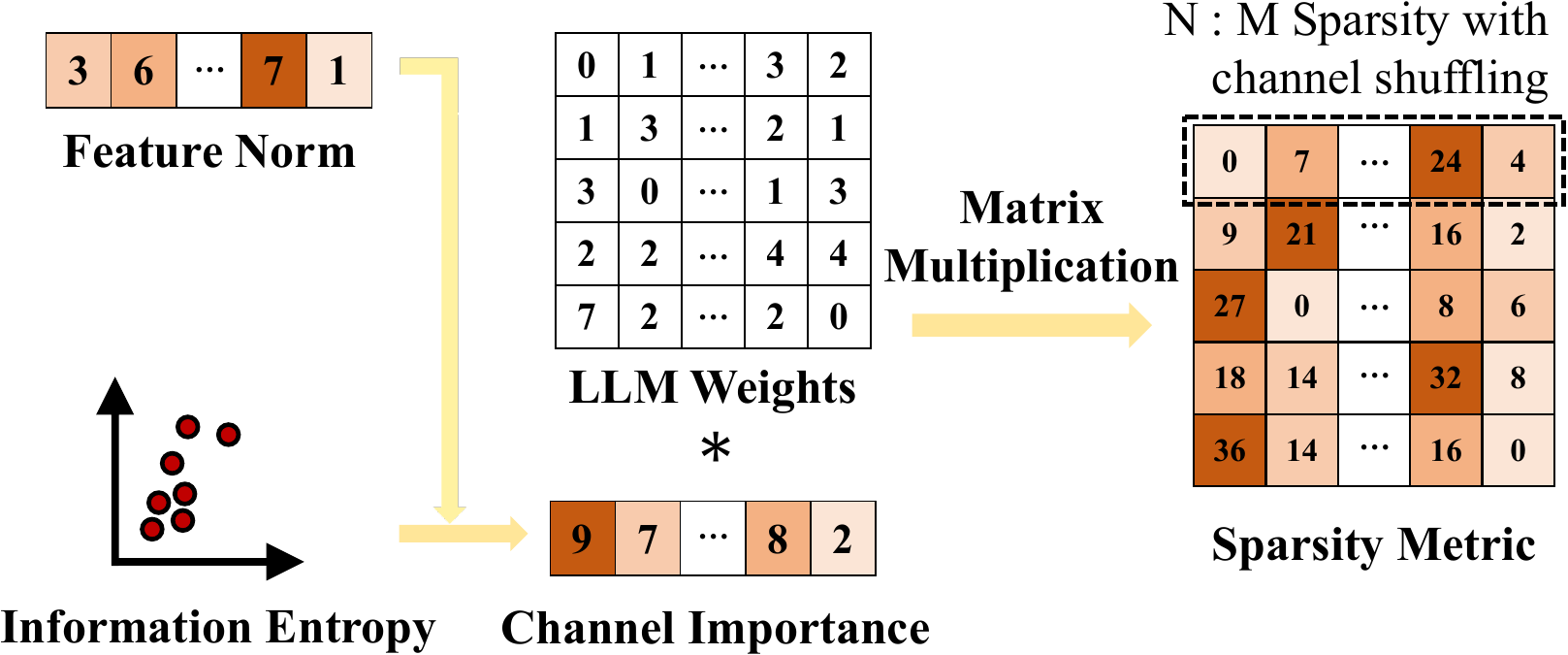} \label{fig:metric_2}}
\caption{Overview of the proposed \name. It first introduces entropy to quantify the information richness within each channel ( intra-channel ) of the input features, and adopts it to enhance the feature norms ( cross-channel ) as a metric to evaluate parameter importance. Furthermore, it proposes Channel Shuffle to reorder the information distribution in LLMs to obtain N:M Sparsity with less information loss.}
\label{fig:overview}
\end{figure}

Large language models (LLMs), such as GPT-3\cite{brown2020language}, LLaMA\cite{touvron2023llama}, Bloom\cite{scao2022bloom}, and others, have recently exhibited outstanding performance across a wide range of tasks, including but not limited to social systems, intelligent conversation, content generation, code creation, etc. However, deploying LLMs poses significant challenges due to their substantial computational demands and high memory requirements. For instance, the most powerful variant, the Bloom model with 176 billion parameters, necessitates a minimum of 350 GB of storage in half-precision (FP16) format. When configured with a batch size of 1 and a sequence length of 128, Bloom-176B inference demands a formidable ensemble of 16 NVIDIA A10 GPUs, each equipped with 24GB memory. Consequently, optimizing these models through compression and pruning has emerged as a critical strategy to reduce parameter counts, thereby decreasing computational overhead and conserving memory resources. 

In order to harness the acceleration and memory reduction potential offered by sparse neural networks, GPU manufacturers have introduced architectural enhancements. Specifically, the invention of Sparse Tensor Core ~\cite{a100, h100, yao2019balanced, cao2019efficient} technology has been pivotal in capitalizing on weight sparsity within Deep Neural Network (DNN) models. This innovation employs a fine-grained structural pruning technique, involving a 2-out-of-4 pruning approach within each partitioned sub-vector. This method effectively balances the computational workload while maximizing parallelism within the dot-product unit.

While there has been substantial research on compressing LLMs using low-precision quantization\cite{xiao2023smoothquant, dettmers2022llm, frantar2022gptq}, relatively little effort has been dedicated to fully exploiting Sparse Tensor Core technology for accelerating LLMs. Some prior work, exemplified by Wanda\cite{sun2023simple}, has proposed the application of a 2-out-of-4 pruning pattern for LLMs. This approach determines channel importance by evaluating input feature norms and weights them against standard parameter magnitudes as pruning metrics. In this studyhe, we introduce an Entropy-based pruning algorithm that builds upon these principles. Our research showcases a remarkable \textbf{1.32} LLaMA perplexity improvement over state-of-the-art techniques and delivers a \textbf{19.6\%-34.8\%} speedup on an A100 GPU, demonstrating the effective and efficient utilization of Sparse Tensor Core hardware.



Our work is grounded in two crucial observations. Firstly, we note that \textbf{the richness of information among channels exhibits significant variation}. Even within the same batch of tokens, the entropy of elements within each channel differs considerably, despite some sharing the same input feature norm. Secondly, we observe that \textbf{channels with close entropy values tend to exhibit relatively concentrated distributions}. These observations naturally inspire us to leverage channel-specific information in order to enhance LLMs inference using $N \colon M$ sparsity.


\textbf{Our proposal} 
We propose entropy-based sparsity (\name ), a novel method to prune LLMs without modifying the remaining weights. Figure \ref{fig:overview} shows the key idea of one-shot \name. 

Firstly, inspired by Observation 1, we introduce a novel metric to assess the importance of weights. This metric employs information entropy to quantify the amount of information within each channel of the hidden state features in LLMs. We enhance the significance of parameter weights and input feature norms by incorporating information entropy as a metric for evaluating parameter importance.

Secondly, we implement a channel shuffling mechanism to ensure a more equitable distribution of information among the channels in the hidden features ( Figure~\ref{fig:shuffle} ). As Observation 2 reveals, the information distribution across channels tends to be highly concentrated, which can impede the accuracy of $N \colon M$ sparsity due to the need to remove N elements from adjacent M elements. Channel shuffling is instrumental in preserving a greater number of elements within information-rich channels, thereby mitigating the impact of parameter pruning on LLMs accuracy.

Lastly, with the robust support of NVIDIA's cuSPARSE\cite{cuSPARSE} and cuSPARSELt\cite{cuSPARSELt} libraries, we have crafted an efficient \name GEMM designed explicitly for LLMs inference and integrated it into FasterTransformer.




\name enables the N:M sparsity of weights for all the matrix multiplications in LLMs, including the LLaMA family, and OPT. The results show that \name outperforms the performance of the state-of-the-art training-free sparsity methods \cite{frantar2023massive,sun2023simple} for LLMs. 
It has also been demonstrated that \name can achieve a 1.24--1.53$\times$ speedup and a 42.64\%--43.52\% memory saving for LLMs with negligible loss in accuracy. 

\section{Inspiration from Observations}
\label{sec:IC}



\begin{figure*}[t]
	\begin{center}
		\includegraphics[width=1.0\linewidth]{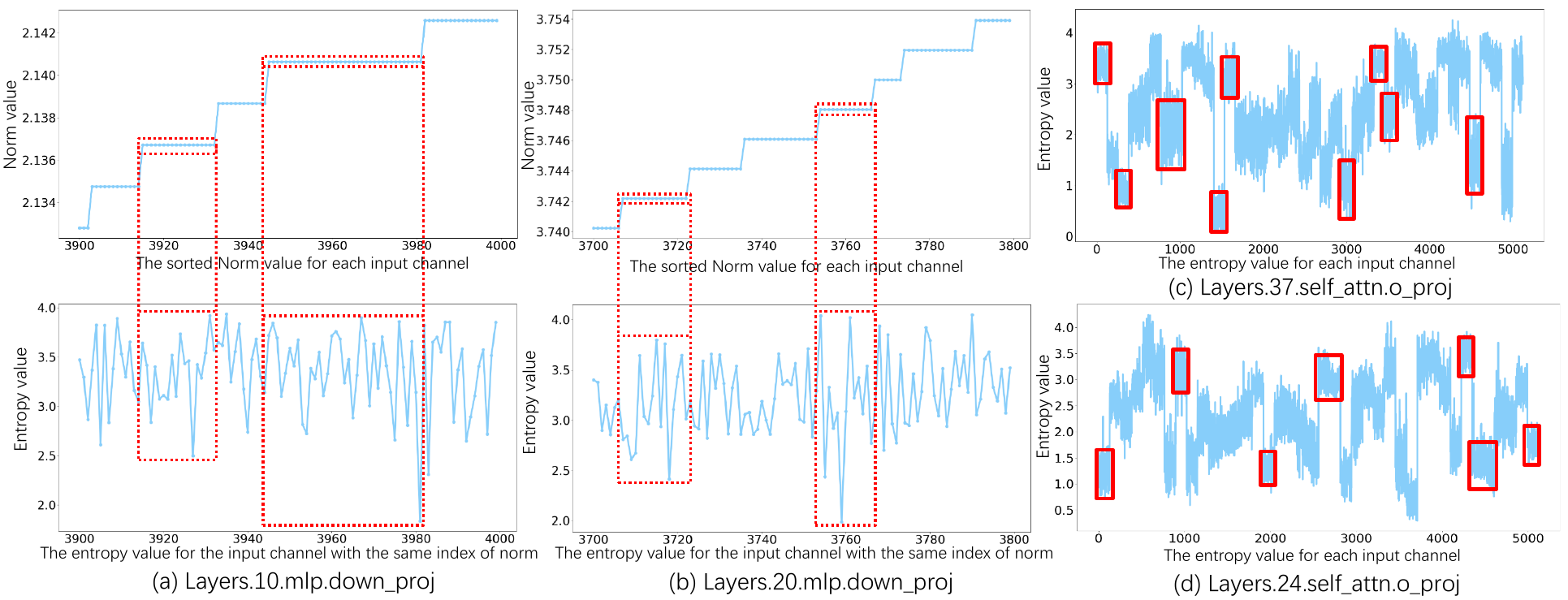}
	\end{center}
	\caption{The visualization of the hidden activations in LLMs. The data for each subfigure comes from the activation of the corresponding layer of LLaMA-13B. For clarity, we only capture the norm and entropy values for the 100 channels after norm sorting in (a) and (b). We show the entropy values of all channels in (c) and (d).}
	\label{fig:observation }
\end{figure*}

It has been found that a small subset of hidden state features (named ``outlier") in LLMs are exceptionally large in magnitude \cite{dettmers2022llm, xiao2023smoothquant}, and these features are important for LLMs compression~\cite{sun2023simple}. Then, we visualize the input activations of linear layers in LLMs and find several key observations about these activations that motivate our method:

\noindent $\bullet$ \textbf{The information richness between channels varies greatly.} A recent work ~\cite{sun2023simple} found that the norm of activation in LLMs can be used to measure channel importance. In addition to the same finding, we also observed that the information entropy between channels also varies greatly. 
To facilitate observation, we first sort the channels according to the norm value and then compare the entropy of each channel feature according to the same index sorted by the norm in Figure~\ref{fig:observation }a and Figure~\ref{fig:observation }b. We find that the entropy of different channels differ considerably, despite some sharing the same input feature norm. The observation above motivates us to enhance evaluation metrics through information richness.

\noindent $\bullet$ \textbf{The entropy values of adjacent channels are relatively close.} As shown in Figure~\ref{fig:observation }c and Figure~\ref{fig:observation }d, channels
with close entropy tend to exhibit relatively concentrated distributions. 
However, N:M sparsity forces the model to prune N values out of M consecutive values in the channel dimension, which makes us inevitably need to prune in M consecutive informative channels and damage the accuracy of LLMs. 
This observation straightforwardly motivates us to shuffle the channels to preserve a greater number of elements within information-rich channels, thereby mitigating the impact of N:M sparsity on accuracy.







\section{Method}

\subsection{Method Overview}

\name proposes a new entropy-based metric to evaluate the parameter importance in LLMs, and introduces channel shuffling to minimize the information loss brought by N:M sparsity. The key advantages of \name include: 
1) Sparse the LLMs without modifying the remaining weights. In contrast to channel-by-channel parameter sparse and update \cite{frantar2023massive}, \name augments the parameter weights with the information richness and the amplitude of the feature as an evaluation metric, and then adopts it to sparse the weights of a layer at once.
2) More fine-grained importance evaluation of hidden state channels. Apart from the global information (channel amplitude), \name introduces entropy to measure the local information of channels (information richness), thereby comprehensively measuring the importance of channels.
3) More flexible sparse mode. Traditional N:M sparsity forces pruning of N out of M consecutive values, \name introduces channel shuffle mechanism, which is more adaptable to the feature information distribution of LLMs and reduces accuracy loss.


\subsection{Information Richness - Entropy}

The observation in Section~\ref{sec:IC} motivates us to enhance the evaluation metrics of LLMs pruning through information richness. Entropy~\cite{shannon1948mathematical} is a key indicator in the field of information theory to measure the amount of information and uncertainty. 
The larger the entropy, the higher the information richness.
Therefore, we introduce entropy to evaluate the channel information of activation for augmenting the standard weight magnitude and channel norm as a novel pruning metric.


Let $X\in \mathbb{R}^{o\times C}$ denote the hidden feature of a fully connected layer in LLMs, where $C$ is the number of channels, and $o$ is the dimension of each channel. To compute the entropy, we first divide it into $K$ different bins and then calculate the probability of an element in the channel falling into each bin. Then, the information richness (entropy) of channel $c$ can be formulated as:

\begin{equation}
 \mathcal{IR}_{c}=-\sum_{k=1}^{K}p_{k}^{c} log\left ( p_{k}^{c}\right ) 
\end{equation}
in which, $p_{k}^{c}$ is the probability of bin $k$ in channel $c$, and ${IR}_{c}\in [0,+\infty )$. 
We set $K$ to 100 empirically, which can achieve good results. 
Information entropy can be used as a good fine-grained metric to evaluate information richness. 
The larger ${IR}_{c}$ value means higher information richness. 

Next, regarding coarse-grained evaluation, we follow \cite{sun2023simple} and adopt the input feature norm to measure the amplitude:

\begin{equation}
 \mathcal{AM}_{c}=\left\| X_{c}\right\|_{2}
\end{equation}
where $\left\| X_{c}\right\|_{2}$ represents the $L^{2}$ norm of the channel $ X_{c}$.

Finally, to comprehensively evaluate the importance of channels and obtain more reasonable weight evaluation metric, we integrated the fine-grained indicator and the coarse-grained indicator above to get the following evaluation metric for pruning redundant weights in LLMs:

\begin{equation}
 \mathcal{\xi} _{cj}=\left| w_{cj}\right|\cdot \left ( \mathcal{IR}_{c}+\alpha\cdot \mathcal{AM}_{c} \right )
\label{metric}
\end{equation}
in which, $w_{cj}$ is the $j$-th element in channel $c$ of the fully connected layer in LLMs, and $\xi _{cj}$ is the final important score of $w_{cj}$ in the sparsity metric.  The larger $\xi _{cj}$ value means higher importance of the element in this layer.

\subsection{Information Reorder - Channel Shuffle}
\label{sec:CS}

\begin{figure}[t]
\centering
\subfloat[Global Naive Shuffle. ]{\includegraphics[width=1.0\linewidth]{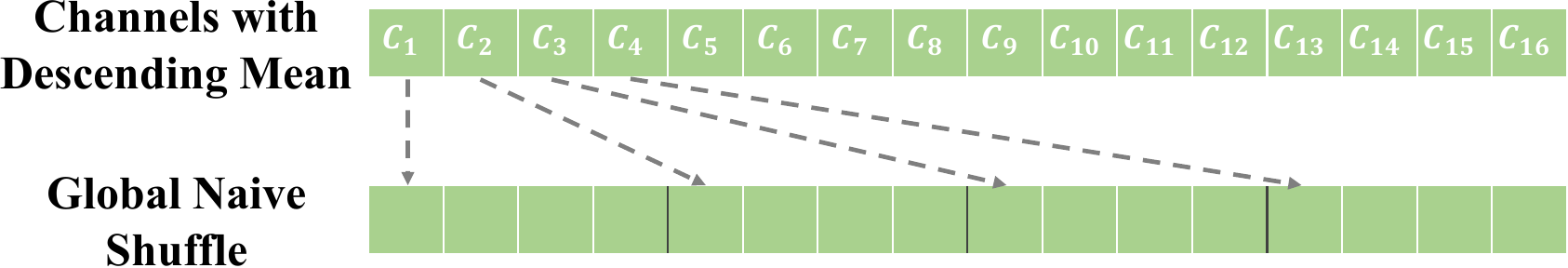} \label{fig:naive}} \\
\vspace{2mm}
\subfloat[Local Block Shuffle.]{\includegraphics[width=1.0\linewidth]{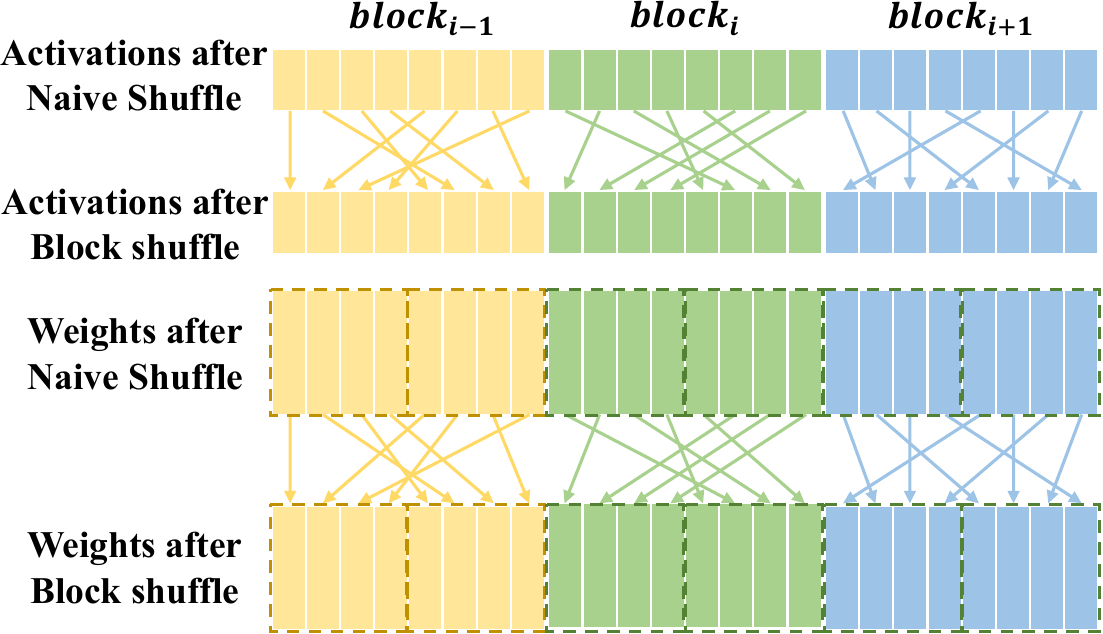} \label{fig:block}}
\caption{Channel Shuffle of \name. Take 2:4 sparsity as an example. \name first sorts the channels \textbf{globally} according to the channel mean of the sparsity metric, and then divides the channels with close mean into different groups, which is \textbf{coarse-grained but faster}.  Then, \name splits the channel into multiple blocks and performs channel shuffle within the blocks, which is slightly slower than the global shuffling but more accurate.}
\label{fig:shuffle}
\end{figure}

Inspired by the observation in Section \ref{sec:IC}, \name implements a channel shuffling mechanism to ensure a more equitable distribution of information among the channels in the hidden features.
By reordering the channel index of the hidden state feature and the layer parameter, \name aims to make the channels with higher information richness distributed more evenly, thus minimizing the information loss caused by N:M sparsity. 

First, the N:M sparsity can be formulated as a constrained optimization problem:

\begin{equation}
\mathcal{O}=\mathop{\min}_{\theta} \frac{1}{2}\left\| Y - W_{N:M}^{\theta }\cdot X \right\|_{F}^{2}
\label{o1}
\end{equation}
in which, $X$ and $Y$ are the input and original output of a fully connected layer, respectively. $\theta$ is the index order of channels, and $W_{N:M}^{\theta }$ is the weight after performing N:M sparsity on W under the current index order. We are committed to finding an optimal channel order $\theta$, which can minimize the output loss caused by M:N sparsity. However, directly optimizing the above problems in LLMs will bring a large computational overhead. Considering that the importance metric in \eqref{metric} contains the information from both weights and activation, we simplify the above problem to minimizing the sparse loss of $\xi_{cj}$:

\begin{equation}
\mathcal{\acute{O}}=\mathop{\max}_{\theta}\sum_{c=1}^{C}(\xi_{cj})_{N:M}^{\theta }
\label{o2}
\end{equation}
in which, $(\xi_{cj})_{N:M}^{\theta }$ is the evaluation metric after N:M sparsity under the channel permutation $\theta$. Compared to \eqref{o1}, there is no need to repeatedly perform matrix multiplication to calculate the feature map $Y$ and the sparse feature map.

Although the optimization problem above has been greatly simplified, performing channel shuffle in LLMs is non-trivial. The large channel size of LLMs results in a big search space, which in turn brings huge computational and time overhead. For a fully connected layer with $C$ channels, there are $C!$ different orderings of channels. For instance, a layer with 1024 channels has a channel ordering of $10^{2640}$. 
In LLMs, the maximum number of channels can reach more than 10,000, which brings huge resistance to obtaining the optimal permutation.

To deal with the issue above, \name introduced the channel shuffle, which consists of two steps: \textit{global naive shuffle} and \textit{local block shuffle}. 

\textbf{Global Naive Shuffle.} To reduce the complexity of channel shuffle in LLMs as much as possible, \name first performs a fast global channel shuffle. For the sparsity metric $\xi \in \mathbb{R}^{o\times C}$, the mean value of each channel is calculated, and based on which the channels are shuffled in descending order. As shown in Figure~\ref{fig:naive}, according to the sparsity pattern ($M$), \name shuffles the channels with close means into different sparse groups. Global naive shuffle can achieve fast coarse-grained information reordering.

\textbf{Local Block Shuffle.} 
To further minimize the information loss caused by N:M sparsity, \name introduces local block shuffle.
First, \name divided the $\xi$ after global naive shuffle into $n$ blocks, and each block contains $m$ channels ($C=m\cdot n$), as shown in Figure~\ref{fig:block}. We use $m = 256$ unless otherwise specified, thus the channel search space is reduced from \textbf{$C!$} to \textbf{$n\cdot256!$}, making the number of unique permutations can be completed in an acceptable amount of time.  
Then, \name performs channel shuffling in each small block by adapting the classic greedy search algorithm~\cite{ji2018tetris, pool2021channel}. 

Combining global naive shuffle and local block shuffle, \name can realize a fast optimization for information distribution and well cope with the challenge of large channel dimensions in LLMs.











\section{Efficient Sparse-GEMM Implementation }

To deploy the proposed method in actual application scenarios, we implemented \name as a sparse engine for efficient LLMs inference. We choose FasterTransformer\cite{FasterTransformer} as the backend and implemented the sparse general matrix multiplication (Sparse-GEMM) of \name for LLMs inference. Taking 2:4 sparsity as an example, the sparse deployment of Sparse-GEMM mainly includes three steps. (1) \name first compresses the sparse weights $ W_{2:4}\in \mathbb{R}^{o\times C}$ into a compressed format, which includes the non-zero weights $ W_{2:4}\in \mathbb{R}^{o\times \frac{C}{2}} $ and the indices of these non-zero data values. (2) With the support of NVIDIA's cuSPARSE and cuSPARSELt, \name searches for the optimal matrix multiplication algorithm according to the shape of each sparse weights tensor in LLMs and saves them. (3) Integrates \name into FasterTransformer for LLMs inference.
Based on the saved optimal matrix multiplication algorithm, LLMs can skip 50\% of matrix multiplication operations and perform faster inference. The experiments in Section \ref{speedup} have shown that such a design can bring \textbf{19.6\%--34.8\%} latency reduction and \textbf{42.64\%--43.52\%} memory saving.

\section{Experiments}
\begin{table*}[t]
\centering
\footnotesize
    \caption{\name 's perplexity performance on LLaMA model family. The results show that \name can outperform state-of-the-art methods by a large margin without updating the remaining weights.
    As for the more constrained and challenging 2:4 sparsity, \name can obtain an 8.26 perplexity for LLaMA-13B, which is \textbf{1.32 better than Wanda and 0.85 better than SparseGPT}.
    }
\label{tab:ppl_llama}
\resizebox{0.7\textwidth}{!}
    {
    \begin{tabular}{cccccccc}
    \toprule
    \textbf{Methods} & \textbf{N:M sparsity} & \textbf{LLaMA-7B} & \textbf{LLaMA-13B} & \textbf{LLaMA-30B} & \textbf{LLaMA-65B}  \\
    \midrule
    FP16 & - & 5.68 & 5.09 & 4.10 & 3.56 \\
    \midrule
    Magnitude & \multirow{4}{*}{2:4} & 42.53 & 18.36 & 7.62 & 7.11 \\
    SparseGPT &  & 11.00 & 9.11 & 7.16 & 6.28 \\
    Wanda &  & 11.53 & 9.58 & 6.90 & 6.25 \\
    \textbf{\name} & & \textbf{10.56} & \textbf{8.26} & \textbf{6.56} & \textbf{5.69}\\
    \midrule
    Magnitude & \multirow{4}{*}{4:8} & 16.83 & 13.86 & 9.11 & 6.35 \\
    SparseGPT &  & 8.61 & 7.40 & 6.17 & 5.38 \\
    Wanda &  & 8.56 & 7.40 & 5.97 & 5.30 \\
    \textbf{\name} & & \textbf{8.29} & \textbf{6.92} & \textbf{5.74} & \textbf{5.09}\\
    \bottomrule
    \end{tabular}}
\end{table*}

\begin{table*}[t]
\centering
\caption{Accuracy of LLaMA under 2:4 sparsity patterns on different Zero-Shot tasks. 
It shows that \name consistently outperforms SparseGPT and Wanda, especially in terms of overall average accuracy across five tasks.
}
\label{tab_acc2}
\resizebox{0.65\linewidth}{!}{
\begin{tabular}{@{}cccccccccc@{}}
\toprule
\textbf{Params} & \textbf{Method} & \textbf{HellaSwag} & \textbf{PiQA} & \textbf{OpenBookQA} & \textbf{SciQ} & \textbf{LogiQA} & \textbf{Avg.} \\ \midrule
\multirow{5}{*}{7B} & FP16 & 56.41 & 78.29 & 28.20 & 89.6 & 21.81 & 54.86 \\
\cmidrule{2-8}
                    & Magnitude & 41.98 & 68.00 & 22.00 & 74.00 & 21.00 & 45.60 \\
                    & Sparse GPT & 42.95 & 70.78 & 19.80 & \textbf{85.00} & \textbf{23.34} & 48.37 \\
                    & Wanda & 41.82 & 70.13 & 21.60 & 83.90 & 21.96 & 47.68 \\
                    & \textbf{\name} & \textbf{43.59} & \textbf{72.03} & \textbf{23.00} & 84.10 & 22.27 & \textbf{49.00} \\
\midrule
\multirow{5}{*}{13B} & FP16 & 59.08 & 78.89 & 30.60 & 93.40 & 26.57 & 59.77 \\
\cmidrule{2-8}
                     & Magnitude & 45.06 & 71.27 & 23.20 & 82.80 & 25.80 & 57.71\\
                     & Sparse GPT & 47.34 & 74.48 & 24.00 & \textbf{88.00} & 21.35 & 51.03 \\
                     & Wanda & 45.99 & 73.55 & \textbf{25.40} & 87.90 & \textbf{23.04}& 51.16 \\
                     & \textbf{\name} &  \textbf{49.40} & \textbf{75.24} & 24.80 & 87.80 &  19.81  & \textbf{51.41} \\
\midrule
\multirow{5}{*}{30B} & FP16 & 62.64 & 81.55 & 29.06 & 92.50 & 28.41 & 58.83 \\
\cmidrule{2-8}
                     & Magnitude & 51.10 & 77.36 & 24.40 & 90.10 & 22.42  & 53.08 \\
                     & Sparse GPT & 52.60 & \textbf{78.40} & 28.20 & 93.30 & 25.96 & 55.69 \\
                     & Wanda & 53.74 & 77.96 & 27.40 & 92.90 & 27.80 & 56.00 \\
                     & \textbf{\name} & \textbf{56.41} & 77.36 & \textbf{28.80} &  \textbf{93.80} & \textbf{29.03} & \textbf{57.08} \\
\midrule
\multirow{5}{*}{65B} & FP16 & 62.64 &  81.55
 & 29.60 & 92.50 & 28.41 
 & 58.94\\
 \cmidrule{2-8}
                     & Magnitude & 57.07 & 77.36 & 30.00 &  90.10 & 23.65  & 55.64 \\
                     & Sparse GPT & 55.23 & 78.40 & 27.60 & 93.30 & 24.42  & 55.79 \\
                     & Wanda & 55.76 & 77.96 & 29.00 &92.90 & \textbf{26.72} & 56.47\\
                     & \textbf{\name} &  \textbf{58.46} & \textbf{78.56} & \textbf{31.60} & \textbf{93.80} & 23.04 & \textbf{57.09} \\
\bottomrule
\end{tabular}
}
\end{table*}

\begin{table}[tb]
\centering
\footnotesize
\vspace{0.2mm}
    \caption{\name 's perplexity performance on OPT models. The results reveal that \name achieves higher performance than Magnitude and Wanda on both 2:4 and 4:8 patterns, which demonstrates the good generalization of \name.}
    \vspace{-0.2cm}
\label{tab:ppl_opt_bloom}
\resizebox{\linewidth}{!}
    {
    \begin{tabular}{cccccc}
    \toprule
    \textbf{Methods}  & 
      \textbf{OPT-6.7b(2:4)} & \textbf{OPT-30b(2:4)} & \textbf{OPT-6.7b(4:8)} & \textbf{OPT-30b(4:8)}  \\
    \midrule
    FP16  & 10.86 & 9.56 & 10.86& 9.56 \\
    \midrule
    Magnitude & 264.14 & 1980.71 & 196.18 & 563.72 \\
    Wanda   & 15.89 & 13.42 & 13.56 & 10.87 \\
    \textbf{\name}  & \textbf{14.90} & \textbf{12.35} & \textbf{13.12} & \textbf{10.75}\\
    \bottomrule
    \end{tabular}}
\end{table}


\subsection{Experimental Environments}
\textbf{Setup.} In our experimental framework, we primarily target the LLaMA model family (LLaMA-7B/13B/30B/65B) and OPT models (OPT-6.7B/30B). To demonstrate the comprehensiveness of E-Sparse, we further extends it to the OPT and BLOOM models. All models are from the HuggingFace Transformers library \cite{wolf2019huggingface}. We choose two SOTA methods as our baselines: SparseGPT and Wanda. Following the one-shot sparsity setting of Wanda, we sample the same 128 sequences from C4 \cite{raffel2020exploring} training data as calibration dataset. All our experiments only need read right on the models without modifying the remaining weights. 
In addition, we demonstrate the real-world inference acceleration of 4:8 and 2:4 sparsity patterns on NVIDIA Ampere Architecture \cite{a100}.

\textbf{Datasets \& Evaluation.} As perplexity is a stable and robust metric to measure the capabilities of LLMs. Importantly, lower perplexity values indicate better model performance. We reported our results on the WikiText \cite{merity2016pointer} validation dataset, based on the perplexity metric. To further demonstrate the efficiency of our method, we also present the zero-shot performance of the pruned networks. Notably, higher values are indicative of superior model performance. Our evaluation rely on the widely-acknowledged EleutherAI LM Harness benchmark \cite{gao2021framework}. The zero-shot evaluation benchmark mainly includes the following datasets: HellaSwag \cite{zellers2019hellaswag}, OpenbookQA \cite{mihaylov2018can}, PiQA \cite{bisk2020piqa}, SciQ \cite{pedersen2020sciq} and LogiQA \cite{liu2020logiqa}.

\begin{table*}[tb]
\centering
\caption{Ablation study on the pruning metric and channel shuffle. Let ${Norm}$ denote the input feature norm (baseline). ${Entropy}$ indicates the information entropy. $GNS$ means the Global Naive Shuffle, and $LBS$ is the Local Block Shuffle. 
The results show that both the proposed entropy strategy and two shuffling methods can bring noteworthy performance gains.}
\label{tab:ablation}
\resizebox{0.75\textwidth}{!}
{ 
\begin{tabular}{@{}cccccccc@{}}
\toprule
 \multicolumn{4}{c}{\textbf{Techniques}} &\multirow{2}{*}{ \textbf{LLaMA-7B}} & \multirow{2}{*}{\textbf{LLaMA-13B}} & \multirow{2}{*}{\textbf{LLaMA-30B}} & \multirow{2}{*}{\textbf{LLaMA-65B}}\\ 
\cmidrule{1-4}
${Norm}$ & ${Entropy}$ & ${GNS}$  &${LBS}$ & & & & \\ 
\midrule
 \Checkmark & \XSolidBrush & \XSolidBrush & \XSolidBrush & 11.53  & 9.58 & 6.90 & 6.25 \\
 \Checkmark & \Checkmark & \XSolidBrush & \XSolidBrush & 11.42  & 8.82 & 6.80 & 6.05\\
 \Checkmark & \Checkmark & \Checkmark & \XSolidBrush & 10.98 & 8.58 & 6.62 & 5.78 \\
  \Checkmark & \Checkmark & \Checkmark & \Checkmark & \textbf{10.56}  &  \textbf{8.26}  & \textbf{6.56} & \textbf{5.69} \\
\bottomrule
\end{tabular}}
\end{table*}



\begin{table*}[tb]
\centering
\footnotesize
\caption{GEMM Speedup of \name after 2:4 sparsity on LLMs. The inputs and weights are all in half-precision (FP16) format, and the latency is evaluated on a single NVIDIA A100 40GB GPU. 
}
\label{tab:latency}
\vspace{-0.2cm}
\resizebox{1\textwidth}{!}
    {
    \begin{tabular}{cccccccc}
    \toprule
    & \textbf{Layer} & \textbf{Input} & \textbf{Weights} & \textbf{Dense GEMM} & \textbf{\name GEMM} & \textbf{Latency Reduction}  \\
    \midrule
    \multirow{4}{*}{Context-stage} & Q/K/V & $16384\times 14336$ & $14336\times 5376$ & 8.452ms & 5.815ms & \textbf{31.2\%} \\
    & Att\_Out & $16384\times 1792$ & $1792\times 14336$ & 3.488ms & 2.540ms & \textbf{27.2\%} \\
    & FFN-1 & $16384\times 14336$ & $14336\times 7168$ & 11.487ms & 8.073ms & \textbf{29.7\%} \\
    & FFN-2 & $16384\times 7168$ & $7168\times 14336$ & 11.478ms & 8.958ms & \textbf{21.9\%} \\
    \midrule
    \multirow{4}{*}{Decoder} & Q/K/V & $16\times 14336$ & $14336\times 5376$ & 0.122ms & 0.098ms & \textbf{19.6\%} \\
    & Att\_Out & $16\times 1792$ & $1792\times 14336$ & 0.046ms & 0.030ms & \textbf{34.8\%} \\
    & FFN-1 & $16\times 14336$ & $14336\times 7168$ & 0.160ms & 0.112ms & \textbf{30.0\%}  \\
    & FFN-2 & $16\times 7168$ & $7168\times 14336$ & 0.158ms & 0.109ms & \textbf{31.0\%}  \\
    \bottomrule
    \end{tabular}}
\end{table*}

\begin{table}[tb]
\centering
\footnotesize
\vspace{0.2mm}
\caption{Memory saving of \name on LLaMA family.}
    \label{tab:memory}
\vspace{-0.2cm}
\resizebox{\linewidth}{!}
    {
    \begin{tabular}{ccccc}
    \toprule
    \textbf{Models}  & 
      \textbf{Dense (FP16)} & \textbf{Sparse (FP16)} & \textbf{Memory Saving}  \\
    \midrule
    LLaMA-7B  & 9.85GB & 5.65GB & \textbf{42.64\%}  \\
    LLaMA-13B  & 19.11GB & 10.89GB & \textbf{43.01\%} \\
    LLaMA-30B  & 47.99GB & 27.17GB & \textbf{43.38\%} \\
    LLaMA-65B  & 96.50GB & 54.50GB & \textbf{43.52\%} \\
    \bottomrule
    \end{tabular}}
\end{table}

\subsection{Pruning Results on LLMs}
To demonstrate the pruning performance of \name, we conduct a series of experiments to evaluate its efficacy across various model sizes within the LLaMA model family.
Similar to Wanda and SparseGPT, we evaluate the perplexity of WikiText validation on structured 4:8 and 2:4 sparsity. As Table \ref{tab:ppl_llama} shows, our \name achieves significant improvements compared with the strong baselines. It is noteworthy that \name does not require weight updates, yet it outperforms the reconstruction-based SparseGPT across all variants within the LLaMA model family. 
At the largest LLaMA-65B, the performance of \name is close to the FP16 baseline. For instance, 4:8 sparsity achieves a perplexity loss of only 1.53 more than FP16. The results indicate that our entropy-based metric and channel shuffle mechanism plays a critical role in N:M sparsity.

To assess the generalization of our method, we conduct experiments on OPT model family, which is one of the most representative LLMs prior to the release of the LLaMA. 
We choose two models of varying sizes, specifically the OPT-6.7B and OPT-30B, for our experiments. According to the result in Table \ref{tab:ppl_opt_bloom}, it is evident that the implementation of \name can lead to a substantial enhancement in WikiText validation. For instance, \name can  achieve a perplexity score of 14.9 at 2:4 sparsity, markedly outperforming Wanda baseline, which registers at 15.89. 

To provide further evidence of our method's performance, we also present results on several ZeroShot tasks for LLaMA under 2:4 sparsity. The comprehensive results have been tabulated in Tab \ref{tab_acc2}. 
It can be observed that our \name consistently exhibits an edge, particularly evident from the superior average accuracy metrics amassed across the quintet of Zero-Shot tasks when compared with other established baseline methods.
\name outperforms Wanda by a margin of 3\% and exceeds SparseGPT by 1\% on average accuracy for LLaMA-7B.
Despite the 2:4 pruning being the most constrained sparsity pattern, our method achieves enhanced performance for all model size on HellaSwag. Additionally, our approach either matches or surpasses the performance of Wanda and SparseGPT on the other four datasets.

\subsection{Ablation Study}

The good performance of \name is mainly attributed to the proposed entropy-based pruning metric and two channel shuffle strategies. 
To validate the effectiveness of these strategies, we conduct a series of ablation studies on LLaMA models in 2:4 sparse pattern.
We take the input feature norm ($Norm$ \cite{sun2023simple}) as the baseline strategy. 

The results are shown in Table~\ref{tab:ablation}. Firstly, it shows that simply introducing \textit{Entropy} to build the pruning metric can bring up to 0.76 perplexity improvement, demonstrating the effectiveness of information entropy on LLM pruning. Then, the introduction of the global naive shuffle and the local block shuffle successively brought the perplexity gains of up to 0.44 and 0.42 respectively, which reveals that $GNS$ and $LBS$ are two complementary channel shuffle strategies.
The results above prove that the three proposed new techniques are efficient and effective.

\subsection{Speedup and Memory Saving}
\label{speedup}

In this section, we show the measured speedup and memory saving of \name integrated into FasterTransformer.

\textbf{Speedup.} With the \name integrated into FasterTransformer, we measure the latency of GEMM in the Context-stage and the Decoder for a batch of 4 and a sequence length of 1024. Due to the lack of support for 4:8 sparsity pattern in NVIDIA Ampere architecture, we only measure the latency of GEMM with 2:4 sparsity on a single A100 40GB GPU. As shown in Table~\ref{tab:latency}, \name is consistently faster than the dense FP16 GEMM baseline, delivering up to 34.8\% latency reduction. It shows that \name can work well on both the context-stage and the decoder in LLMs.

\textbf{Memory Saving.} In Table \ref{tab:memory}, we give the memory saving brought by \name on LLaMA family. The results reveal that it can save 42.64\%--43.52\% memory usage on LLaMA models. We can also see a trend that the larger the model, the more significant the memory saving.

\section{Related Work}

 \textbf{Traditional Network Pruning.} 
 Network pruning was proposed to remove redundant parts of the DNN models, thereby reducing the computational and memory demands of neural networks without accuracy loss \cite{liu2018rethinking,louizos2017learning, han2015deep, hassibi1993optimal}. 
 Traditional network pruning techniques usually fall into two primary categories: unstructured pruning~\cite{hassibi1993optimal, han2015learning, han2015deep} and structured pruning~\cite{li2016pruning, luo2017thinet, liu2017learning, li2020weight, li2022weight, ding2021resrep, li2021boosting, xia2022structured}.  
 Unstructured pruning methods \cite{han2015deep, han2015learning} aim to iteratively prune unimportant connections whose absolute weights are smaller than a given threshold, which achieves good performance on parameter compression. 
 However, such kind of methods are implementation-unfriendly.
  Structured pruning methods \cite{li2016pruning, luo2017thinet, liu2019metapruning} prune or sparse entire parts of the network (e.g., channels, blocks) instead of individual weights, thus require less specialized libraries to achieve inference speedup.
 A common feature of the traditional pruning techniques mentioned above is that the pruned network usually needs to be retrained to recover the accuracy loss, which hinders their application on LLMs that consume huge training resources.

\textbf{N:M Sparsity.} N:M sparsity~\cite{mishra2021accelerating,pool2021channel,akiva2022searching,zhou2021learning} is a kind of special pruning technique that introduces an intermediate sparsity pattern between unstructured and structured pruning, called semi-structured sparsity. N:M sparsity aims to prune N out of every M consecutive parameters, rather than pruning individual weights or entire channels/blocks. The appeal of N:M sparsity is its ability to reason for specific hardware architectures (such as NVIDIA Ampere\cite{pool2020accelerating}), enabling efficient computation. \cite{akiva2022searching} suggests a Neural Architecture Search (NAS) strategy to sparse both activations and weights throughout the network. \cite{zhou2021learning} defines a metric, Sparse Architecture Divergence (SAD) to learn N:M sparse neural networks. However, these are only designed for CNNs or small models, and how to design efficient N:M sparsity for LLMs has been rarely studied.

\textbf{Pruning for LLMs.} Due to the massive size and computational costs of large language models, training-based pruning methods \cite{ma2023llm, xia2023sheared, singh2023exploiting} will bring a large overhead. So existing popular solutions aim at post-training pruning strategy\cite{frantar2023massive, sun2023simple}. Such methods only need a small number of calibration data to prune the pre-trained LLMs models, which is suitable for rapid deployment.
SparseGPT\cite{frantar2023massive} develops a layer-wise weight update for LLMs via an approximate second-order Hessian. This schema is iteratively executed between weight pruning and weight update at each layer, which is computationally expensive.
Wanda\cite{sun2023simple} presents to remove the insignificant weights based on the magnitude and norm of corresponding input activations, without updating the remaining weights.
Our work further proposes a new metric based on the information richness and designs an effective search strategy for N:M sparsity.

\section{Conclusion}

In this paper, we propose a novel entropy-based pruning method, called \name, to carry out N:M sparsity on LLMs in a one-shot manner. 
The design of our pruning metric is based on the observation of the information richness of hidden state channels and relatively concentrated distributions of information-rich channels.
Extensive experiments show the superior performance of our proposal against existing LLMs pruning methods.

\section{Limitations}

Beyond NLP tasks, the applicability of \name to other tasks (including computer vision or speech recognition), remains to be tested. For fair comparison with other methods, we only conducted experiments on public datasets with limited sentence lengths. In addition, the combined optimization of \name and other orthogonal methods (quantization or distillation) has not yet been studied.

\bibliography{main}

\begin{thebibliography}{48}
\expandafter\ifx\csname natexlab\endcsname\relax\def\natexlab#1{#1}\fi

\bibitem[{a10(2020)}]{a100}
 2020.
\newblock {NVIDIA A100 Tensor Core GPU Architecture}.
\newblock \url{https://images.nvidia.com/aem-dam/en-zz/Solutions/data-center/nvidia-ampere-architecture-whitepaper.pdf}.

\bibitem[{cuS(2023{\natexlab{a}})}]{cuSPARSE}
 2023{\natexlab{a}}.
\newblock {cuSPARSE}.
\newblock \url{https://docs.nvidia.com/cuda/cusparselt/index.html}.

\bibitem[{cuS(2023{\natexlab{b}})}]{cuSPARSELt}
 2023{\natexlab{b}}.
\newblock {cuSPARSELt}.
\newblock \url{https://docs.nvidia.com/cuda/cusparselt/index.html}.

\bibitem[{Fas(2023)}]{FasterTransformer}
 2023.
\newblock {FasterTransformer}.
\newblock \url{https://github.com/NVIDIA/FasterTransformer/tree/release/v5.3_tag}.

\bibitem[{h10(2023)}]{h100}
 2023.
\newblock {NVIDIA H100 Tensor Core GPU Architecture}.
\newblock \url{https://resources.nvidia.com/en-us-tensor-core}.

\bibitem[{Akiva-Hochman et~al.(2022)Akiva-Hochman, Finder, Turek, and Treister}]{akiva2022searching}
Ruth Akiva-Hochman, Shahaf~E Finder, Javier~S Turek, and Eran Treister. 2022.
\newblock Searching for n: M fine-grained sparsity of weights and activations in neural networks.
\newblock In \emph{European Conference on Computer Vision}, pages 130--143. Springer.

\bibitem[{Bisk et~al.(2020)Bisk, Zellers, Gao, Choi et~al.}]{bisk2020piqa}
Yonatan Bisk, Rowan Zellers, Jianfeng Gao, Yejin Choi, et~al. 2020.
\newblock Piqa: Reasoning about physical commonsense in natural language.
\newblock In \emph{Proceedings of the AAAI conference on artificial intelligence}, volume~34, pages 7432--7439.

\bibitem[{Brown et~al.(2020)Brown, Mann, Ryder, Subbiah, Kaplan, Dhariwal, Neelakantan, Shyam, Sastry, Askell et~al.}]{brown2020language}
Tom Brown, Benjamin Mann, Nick Ryder, Melanie Subbiah, Jared~D Kaplan, Prafulla Dhariwal, Arvind Neelakantan, Pranav Shyam, Girish Sastry, Amanda Askell, et~al. 2020.
\newblock Language models are few-shot learners.
\newblock \emph{Advances in neural information processing systems}, 33:1877--1901.

\bibitem[{Cao et~al.(2019)Cao, Zhang, Yao, Xiao, Nie, Zhan, Liu, Wu, and Zhang}]{cao2019efficient}
Shijie Cao, Chen Zhang, Zhuliang Yao, Wencong Xiao, Lanshun Nie, Dechen Zhan, Yunxin Liu, Ming Wu, and Lintao Zhang. 2019.
\newblock Efficient and effective sparse lstm on fpga with bank-balanced sparsity.
\newblock In \emph{Proceedings of the 2019 ACM/SIGDA International Symposium on Field-Programmable Gate Arrays}, pages 63--72.

\bibitem[{Dettmers et~al.(2022)Dettmers, Lewis, Belkada, and Zettlemoyer}]{dettmers2022llm}
Tim Dettmers, Mike Lewis, Younes Belkada, and Luke Zettlemoyer. 2022.
\newblock Llm. int8 (): 8-bit matrix multiplication for transformers at scale.
\newblock \emph{arXiv preprint arXiv:2208.07339}.

\bibitem[{Ding et~al.(2021)Ding, Hao, Tan, Liu, Han, Guo, and Ding}]{ding2021resrep}
Xiaohan Ding, Tianxiang Hao, Jianchao Tan, Ji~Liu, Jungong Han, Yuchen Guo, and Guiguang Ding. 2021.
\newblock Resrep: Lossless cnn pruning via decoupling remembering and forgetting.
\newblock In \emph{Proceedings of the IEEE/CVF International Conference on Computer Vision}, pages 4510--4520.

\bibitem[{Frantar and Alistarh(2023)}]{frantar2023massive}
Elias Frantar and Dan Alistarh. 2023.
\newblock Massive language models can be accurately pruned in one-shot.
\newblock \emph{arXiv preprint arXiv:2301.00774}.

\bibitem[{Frantar et~al.(2022)Frantar, Ashkboos, Hoefler, and Alistarh}]{frantar2022gptq}
Elias Frantar, Saleh Ashkboos, Torsten Hoefler, and Dan Alistarh. 2022.
\newblock Gptq: Accurate post-training quantization for generative pre-trained transformers.
\newblock \emph{arXiv preprint arXiv:2210.17323}.

\bibitem[{Gao et~al.(2021)Gao, Tow, Biderman, Black, DiPofi, Foster, Golding, Hsu, McDonell, Muennighoff et~al.}]{gao2021framework}
Leo Gao, Jonathan Tow, Stella Biderman, Sid Black, Anthony DiPofi, Charles Foster, Laurence Golding, Jeffrey Hsu, Kyle McDonell, Niklas Muennighoff, et~al. 2021.
\newblock A framework for few-shot language model evaluation.
\newblock \emph{Version v0. 0.1. Sept}.

\bibitem[{Han et~al.(2016)Han, Mao, and Dally}]{han2015deep}
Song Han, Huizi Mao, and William~J Dally. 2016.
\newblock Deep compression: Compressing deep neural networks with pruning, trained quantization and huffman coding.
\newblock In \emph{Proceedings of International Conference on Learning Representations (ICLR)}.

\bibitem[{Han et~al.(2015)Han, Pool, Tran, and Dally}]{han2015learning}
Song Han, Jeff Pool, John Tran, and William Dally. 2015.
\newblock Learning both weights and connections for efficient neural network.
\newblock In \emph{Advances in Neural Information Processing Systems (NeurIPS)}, pages 1135--1143.

\bibitem[{Hassibi et~al.(1993)Hassibi, Stork, and Wolff}]{hassibi1993optimal}
Babak Hassibi, David~G Stork, and Gregory~J Wolff. 1993.
\newblock Optimal brain surgeon and general network pruning.
\newblock In \emph{IEEE international conference on neural networks}, pages 293--299. IEEE.

\bibitem[{Ji et~al.(2018)Ji, Liang, Deng, Zhang, Zhang, and Xie}]{ji2018tetris}
Yu~Ji, Ling Liang, Lei Deng, Youyang Zhang, Youhui Zhang, and Yuan Xie. 2018.
\newblock Tetris: Tile-matching the tremendous irregular sparsity.
\newblock \emph{Advances in neural information processing systems}, 31.

\bibitem[{Li et~al.(2017)Li, Kadav, Durdanovic, Samet, and Graf}]{li2016pruning}
Hao Li, Asim Kadav, Igor Durdanovic, Hanan Samet, and Hans~Peter Graf. 2017.
\newblock Pruning filters for efficient convnets.
\newblock In \emph{Proceedings of International Conference on Learning Representations (ICLR)}.

\bibitem[{Li et~al.(2022)Li, Liu, Wu, Yao, Zhang, Zhang, and Yin}]{li2022weight}
Yun Li, Zechun Liu, Weiqun Wu, Haotian Yao, Xiangyu Zhang, Chi Zhang, and Baoqun Yin. 2022.
\newblock Weight-dependent gates for network pruning.
\newblock \emph{IEEE Transactions on Circuits and Systems for Video Technology}, 32(10):6941--6954.

\bibitem[{Li et~al.(2020)Li, Wu, Liu, Zhang, Zhang, Yao, and Yin}]{li2020weight}
Yun Li, Weiqun Wu, Zechun Liu, Chi Zhang, Xiangyu Zhang, Haotian Yao, and Baoqun Yin. 2020.
\newblock Weight-dependent gates for differentiable neural network pruning.
\newblock In \emph{European Conference on Computer Vision Workshops}, pages 23--37. Springer.

\bibitem[{Li et~al.(2021)Li, Zhang, Han, Zhang, Yin, Liu, and Xu}]{li2021boosting}
Yun Li, Chen Zhang, Shihao Han, Li~Lyna Zhang, Baoqun Yin, Yunxin Liu, and Mengwei Xu. 2021.
\newblock Boosting mobile cnn inference through semantic memory.
\newblock In \emph{Proceedings of the 29th ACM International Conference on Multimedia}, pages 2362--2371.

\bibitem[{Liu et~al.(2020)Liu, Cui, Liu, Huang, Wang, and Zhang}]{liu2020logiqa}
Jian Liu, Leyang Cui, Hanmeng Liu, Dandan Huang, Yile Wang, and Yue Zhang. 2020.
\newblock Logiqa: A challenge dataset for machine reading comprehension with logical reasoning.
\newblock \emph{arXiv preprint arXiv:2007.08124}.

\bibitem[{Liu et~al.(2019)Liu, Mu, Zhang, Guo, Yang, Cheng, and Sun}]{liu2019metapruning}
Zechun Liu, Haoyuan Mu, Xiangyu Zhang, Zichao Guo, Xin Yang, Kwang-Ting Cheng, and Jian Sun. 2019.
\newblock Metapruning: Meta learning for automatic neural network channel pruning.
\newblock In \emph{Proceedings of the IEEE international conference on computer vision (ICCV)}, pages 3296--3305.

\bibitem[{Liu et~al.(2017)Liu, Li, Shen, Huang, Yan, and Zhang}]{liu2017learning}
Zhuang Liu, Jianguo Li, Zhiqiang Shen, Gao Huang, Shoumeng Yan, and Changshui Zhang. 2017.
\newblock Learning efficient convolutional networks through network slimming.
\newblock In \emph{Proceedings of the IEEE international conference on computer vision}, pages 2736--2744.

\bibitem[{Liu et~al.(2018)Liu, Sun, Zhou, Huang, and Darrell}]{liu2018rethinking}
Zhuang Liu, Mingjie Sun, Tinghui Zhou, Gao Huang, and Trevor Darrell. 2018.
\newblock Rethinking the value of network pruning.
\newblock \emph{arXiv preprint arXiv:1810.05270}.

\bibitem[{Louizos et~al.(2017)Louizos, Welling, and Kingma}]{louizos2017learning}
Christos Louizos, Max Welling, and Diederik~P Kingma. 2017.
\newblock Learning sparse neural networks through $ l\_0 $ regularization.
\newblock \emph{arXiv preprint arXiv:1712.01312}.

\bibitem[{Luo et~al.(2017)Luo, Wu, and Lin}]{luo2017thinet}
Jian-Hao Luo, Jianxin Wu, and Weiyao Lin. 2017.
\newblock Thinet: A filter level pruning method for deep neural network compression.
\newblock In \emph{Proceedings of the IEEE international conference on computer vision (ICCV)}, pages 5058--5066.

\bibitem[{Ma et~al.(2023)Ma, Fang, and Wang}]{ma2023llm}
Xinyin Ma, Gongfan Fang, and Xinchao Wang. 2023.
\newblock Llm-pruner: On the structural pruning of large language models.
\newblock \emph{arXiv preprint arXiv:2305.11627}.

\bibitem[{Merity et~al.(2016)Merity, Xiong, Bradbury, and Socher}]{merity2016pointer}
Stephen Merity, Caiming Xiong, James Bradbury, and Richard Socher. 2016.
\newblock Pointer sentinel mixture models.
\newblock \emph{arXiv preprint arXiv:1609.07843}.

\bibitem[{Mihaylov et~al.(2018)Mihaylov, Clark, Khot, and Sabharwal}]{mihaylov2018can}
Todor Mihaylov, Peter Clark, Tushar Khot, and Ashish Sabharwal. 2018.
\newblock Can a suit of armor conduct electricity? a new dataset for open book question answering.
\newblock \emph{arXiv preprint arXiv:1809.02789}.

\bibitem[{Mishra et~al.(2021)Mishra, Latorre, Pool, Stosic, Stosic, Venkatesh, Yu, and Micikevicius}]{mishra2021accelerating}
Asit Mishra, Jorge~Albericio Latorre, Jeff Pool, Darko Stosic, Dusan Stosic, Ganesh Venkatesh, Chong Yu, and Paulius Micikevicius. 2021.
\newblock Accelerating sparse deep neural networks.
\newblock \emph{arXiv preprint arXiv:2104.08378}.

\bibitem[{Pedersen et~al.(2020)Pedersen, Otokiak, Koonoo, Milton, Maktar, Anaviapik, Milton, Porter, Scott, Newman et~al.}]{pedersen2020sciq}
C~Pedersen, M~Otokiak, I~Koonoo, J~Milton, E~Maktar, A~Anaviapik, M~Milton, G~Porter, A~Scott, C~Newman, et~al. 2020.
\newblock Sciq: an invitation and recommendations to combine science and inuit qaujimajatuqangit for meaningful engagement of inuit communities in research.
\newblock \emph{Arctic Science}, 6(3):326--339.

\bibitem[{Pool(2020)}]{pool2020accelerating}
Jeff Pool. 2020.
\newblock Accelerating sparsity in the nvidia ampere architecture.
\newblock \emph{GTC 2020}.

\bibitem[{Pool and Yu(2021)}]{pool2021channel}
Jeff Pool and Chong Yu. 2021.
\newblock Channel permutations for n: M sparsity.
\newblock \emph{Advances in neural information processing systems}, 34:13316--13327.

\bibitem[{Raffel et~al.(2020)Raffel, Shazeer, Roberts, Lee, Narang, Matena, Zhou, Li, and Liu}]{raffel2020exploring}
Colin Raffel, Noam Shazeer, Adam Roberts, Katherine Lee, Sharan Narang, Michael Matena, Yanqi Zhou, Wei Li, and Peter~J Liu. 2020.
\newblock Exploring the limits of transfer learning with a unified text-to-text transformer.
\newblock \emph{The Journal of Machine Learning Research}, 21(1):5485--5551.

\bibitem[{Scao et~al.(2022)Scao, Fan, Akiki, Pavlick, Ili{\'c}, Hesslow, Castagn{\'e}, Luccioni, Yvon, Gall{\'e} et~al.}]{scao2022bloom}
Teven~Le Scao, Angela Fan, Christopher Akiki, Ellie Pavlick, Suzana Ili{\'c}, Daniel Hesslow, Roman Castagn{\'e}, Alexandra~Sasha Luccioni, Fran{\c{c}}ois Yvon, Matthias Gall{\'e}, et~al. 2022.
\newblock Bloom: A 176b-parameter open-access multilingual language model.
\newblock \emph{arXiv preprint arXiv:2211.05100}.

\bibitem[{Shannon(1948)}]{shannon1948mathematical}
Claude~Elwood Shannon. 1948.
\newblock A mathematical theory of communication.
\newblock \emph{Bell system technical journal}, 27(3):379--423.

\bibitem[{Singh and Bhatele(2023)}]{singh2023exploiting}
Siddharth Singh and Abhinav Bhatele. 2023.
\newblock Exploiting sparsity in pruned neural networks to optimize large model training.
\newblock \emph{arXiv preprint arXiv:2302.05045}.

\bibitem[{Sun et~al.(2023)Sun, Liu, Bair, and Kolter}]{sun2023simple}
Mingjie Sun, Zhuang Liu, Anna Bair, and J~Zico Kolter. 2023.
\newblock A simple and effective pruning approach for large language models.
\newblock \emph{arXiv preprint arXiv:2306.11695}.

\bibitem[{Touvron et~al.(2023)Touvron, Lavril, Izacard, Martinet, Lachaux, Lacroix, Rozi{\`e}re, Goyal, Hambro, Azhar et~al.}]{touvron2023llama}
Hugo Touvron, Thibaut Lavril, Gautier Izacard, Xavier Martinet, Marie-Anne Lachaux, Timoth{\'e}e Lacroix, Baptiste Rozi{\`e}re, Naman Goyal, Eric Hambro, Faisal Azhar, et~al. 2023.
\newblock Llama: Open and efficient foundation language models.
\newblock \emph{arXiv preprint arXiv:2302.13971}.

\bibitem[{Wolf et~al.(2019)Wolf, Debut, Sanh, Chaumond, Delangue, Moi, Cistac, Rault, Louf, Funtowicz et~al.}]{wolf2019huggingface}
Thomas Wolf, Lysandre Debut, Victor Sanh, Julien Chaumond, Clement Delangue, Anthony Moi, Pierric Cistac, Tim Rault, R{\'e}mi Louf, Morgan Funtowicz, et~al. 2019.
\newblock Huggingface's transformers: State-of-the-art natural language processing.
\newblock \emph{arXiv preprint arXiv:1910.03771}.

\bibitem[{Xia et~al.(2023)Xia, Gao, Zeng, and Chen}]{xia2023sheared}
Mengzhou Xia, Tianyu Gao, Zhiyuan Zeng, and Danqi Chen. 2023.
\newblock Sheared llama: Accelerating language model pre-training via structured pruning.
\newblock \emph{arXiv:2310.06694}.

\bibitem[{Xia et~al.(2022)Xia, Zhong, and Chen}]{xia2022structured}
Mengzhou Xia, Zexuan Zhong, and Danqi Chen. 2022.
\newblock Structured pruning learns compact and accurate models.
\newblock In \emph{Proceedings of the 60th Annual Meeting of the Association for Computational Linguistics (Volume 1: Long Papers)}, pages 1513--1528.

\bibitem[{Xiao et~al.(2023)Xiao, Lin, Seznec, Wu, Demouth, and Han}]{xiao2023smoothquant}
Guangxuan Xiao, Ji~Lin, Mickael Seznec, Hao Wu, Julien Demouth, and Song Han. 2023.
\newblock Smoothquant: Accurate and efficient post-training quantization for large language models.
\newblock In \emph{International Conference on Machine Learning}, pages 38087--38099. PMLR.

\bibitem[{Yao et~al.(2019)Yao, Cao, Xiao, Zhang, and Nie}]{yao2019balanced}
Zhuliang Yao, Shijie Cao, Wencong Xiao, Chen Zhang, and Lanshun Nie. 2019.
\newblock Balanced sparsity for efficient dnn inference on gpu.
\newblock In \emph{Proceedings of the AAAI conference on artificial intelligence}, volume~33, pages 5676--5683.

\bibitem[{Zellers et~al.(2019)Zellers, Holtzman, Bisk, Farhadi, and Choi}]{zellers2019hellaswag}
Rowan Zellers, Ari Holtzman, Yonatan Bisk, Ali Farhadi, and Yejin Choi. 2019.
\newblock Hellaswag: Can a machine really finish your sentence?
\newblock \emph{arXiv preprint arXiv:1905.07830}.

\bibitem[{Zhou et~al.(2021)Zhou, Ma, Zhu, Liu, Zhang, Yuan, Sun, and Li}]{zhou2021learning}
Aojun Zhou, Yukun Ma, Junnan Zhu, Jianbo Liu, Zhijie Zhang, Kun Yuan, Wenxiu Sun, and Hongsheng Li. 2021.
\newblock Learning n: m fine-grained structured sparse neural networks from scratch.
\newblock \emph{arXiv preprint arXiv:2102.04010}.

\end{thebibliography}

\newpage

\appendix

\section{Appendix}

\begin{algorithm}[tbh]
\caption{Channel Shuffle}
\label{alg:1}
\textbf{Input}: Sparsity Metric: $\xi$; Sparsity pattern: $M$ and $N$. \\
\textbf{Output}: Permutations: $\theta$.\\
\begin{algorithmic}[1]
\STATE Initialize the permutations with a global shuffle: $\theta$ = Global\_Naive\_Shuffle($\xi$, M) \\
\STATE Reorder the $\xi$ according to $\theta$. \\
\STATE Divide $\xi$ into $N$ blocks in the channel dimension $\left\{ \xi_{1}, \xi_{2},\cdots ,\xi_{N} \right\}$.
\FOR{$\xi_{n}$ in $\xi$}
\WHILE {True}
\STATE res = search\_permutations\_with\_largest\_ improvement( $\xi_{n}$, M, N ) 
\IF{res is None} 
\STATE break
\ENDIF 
\STATE i, j = res[0], res[1]
\STATE Swap channel $i$ and channel $j$ in $\xi_{n}$. \\
\STATE Update $\theta$.
\ENDWHILE
\ENDFOR
\end{algorithmic}
\end{algorithm}

\end{document}